\documentclass{llncs}

\usepackage{latexsym}
\usepackage{psfig}
\usepackage{url}

\newcommand{\eplc}{\mathit{PS}^+}

\title{{\em aspps} --- an implementation of answer-set programming with
propositional schemata}

\author{Deborah East and Miros\l aw Truszczy\'nski}
\institute{ Department of Computer Science\\ University of Kentucky\\
Lexington KY 40506-0046, USA}

\begin{document}

\date{}
\maketitle
\begin{abstract}
We present an implementation of an answer-set programming
paradigm, called {\em aspps} (short for answer-set  programming with 
propositional schemata). 
The system {\em aspps} is designed to process $\eplc$-theories. It
consists of two basic modules. The first module, {\em psgrnd}, 
grounds an $\eplc$-theory. The second module, referred to as {\em aspps}, 
is a solver. It computes models of ground $\eplc$-theories. 
\end{abstract}

\section{Introduction}
The most advanced answer-set programming systems are, at present, {\em
smodels} \cite{ns00} and {\em dlv} \cite{elmps98}. They are based on 
the formalisms of logic programming with stable-model semantics and 
disjunctive logic programming with answer-set semantics, respectively. 
We present an implementation of an answer-set programming system, {\em 
aspps} (short for answer-set programming with propositional schemata). 
It is based on the {\em extended logic of propositional schemata with 
closed world assumption} that we denote by $\eplc$. We introduced this
logic in \cite{et01a}.
 
A theory in the logic $\eplc$ is a pair $(D,P)$, where $D$ is a
collection of ground atoms representing a {\em problem instance}
(input data), and $P$ is a {\em program} --- a collection of 
$\eplc$-clauses (encoding of a problem to solve). 
The meaning of a $\eplc$-theory $T=(D,P)$ is given by a {\em
family} of $\eplc$-models \cite{et01a}. Each model in this family
represents a solution to a problem encoded by $P$ for data instance
$D$.

The system {\em aspps} is designed to process $\eplc$-theories. It 
consists of two basic programs. The first of them, 
{\em psgrnd}, grounds a $\eplc$-theory. That is, it produces a ground 
(propositional) theory extended by a number of special constructs.
These constructs help model cardinality constraints on sets. The second 
program, referred to as {\em aspps}, is a solver. It computes models of 
grounded $\eplc$-theories. It is designed along the lines of a standard 
Davis-Putnam algorithm for satisfiability checking. Both {\em psgrnd}
and {\em aspps}, examples of $\eplc$-programs and the corresponding 
performance results are available at \url{http://www.cs.uky.edu/ai/aspps/}.             

\section{$\eplc$-theories}

A {\em $\eplc$-theory} is a pair $(D,P)$, where $D$ is a collection of
ground atoms and $P$ is a collection of $\eplc$-clauses. 
Atoms in $D$ represent input data (an instance of a problem). In our 
implementation these atoms may be stored in one or more {\em data} files. 
The set of $\eplc$-clauses models the constraints (specification) of
the problem. In our implementation, all the $\eplc$-clauses in $P$ are 
stored in a single {\em rule} file.

All statements in data and rule files must end with a period (.). 
Clauses may be split across several
lines. Blank lines can be used in data and rule files to improve 
readability. Comments may be used too. They begin with `{\bf \%}' and 
continue to the end of the line. 

\smallskip
\noindent
{\bf Data files.} 
Each ground atom in a data file must 
be given on a single line. 
Constant symbols may be used
as arguments of ground atoms. In such cases, these constant symbols must 
be specified at the command line (see Section \ref{exec}). Examples of
ground atoms are given below:

\vspace*{-0.1in}
\begin{quote}
$\mathit{vtx}(2).$\\
$\mathit{vtx}(3).$\\
$\mathit{size}(k).$
\end{quote}

\vspace*{-0.1in}
\noindent
A set of ground atoms of the form $\{p(m),p(m+1),\ldots,p(n)\}$, where
$m$ and $n$ are non-negative integers or integer constants specified
at the command line, can be represented in a data file as
`$p[m..n]$.'.
Thus, the two ground atoms
$\mathit{vtx}(2)$ and $\mathit{vtx}(3)$ can be
specified as `$\mathit{vtx}[1..3]$.'.

Predicates used by ground atoms in data files are called {\em data
predicates}.

\smallskip
\noindent
{\bf Rule files.} The rule file of a $\eplc$-theory consists of two 
parts. In the
first one, the {\em preamble}, we  declare all {\em program} predicates,
that is, predicates that are not used in data files. We also declare
types of all variables that will be used in the rule files. Typing
of variables simplifies the implementation of the grounding program
{\em psgrnd} and facilitates error checking.

Arguments of each program predicate are typed by unary {\em data}
predicates (the idea is that when grounding, each argument can only be
replaced by an element of an extension of the corresponding unary
data predicate as specified by the data files). A program predicate $q$ 
with $n$ arguments of types $dp_1,\ldots, dp_n$, where all $dp_i$ are 
data predicates, is declared in one of the following two ways: 

\vspace*{-0.1in}
\begin{quote}
$\mathit{pred}\ q(dp_1, \dots, dp_n).$\\
$\mathit{pred}\ q(dp_1, \dots, dp_n): dp_m.$
\end{quote}

\vspace*{-0.1in}
\noindent
In the second statement, the $n$-ary data predicate $dp_m$ further
restricts the extension of $q$ --- it must be a subset of the extension
of $dp_m$ (as specified by the data files). 

Variable declarations begin with the keyword $\mathit{var}$. It is
followed by the {\em unary} data predicate name and a list of
alpha-numeric strings serving as variable names (they must start with 
a letter). 
Thus, to declare two variables $X$ and $Y$ of type $dp$, where $dp$
is a unary data predicate we write:
\begin{quote}
$\mathit{var}\ dp\ X, Y$.
\end{quote}

The implementation allows for {\em predefined} predicates and function 
symbols such as the equality operator $==$, arithmetic comparators 
$<=$, $>=$, $<$ and $>$, and arithmetic operations $+$, $-$, $*$ ,$/$,
$\mathit{abs}()$ (absolute value), $\mathit{mod}(N,b)$, $\mathit{max}
(X,Y)$ and $min(X,Y)$. We assign to these symbols their standard 
interpretation. However, we emphasize that the domains are restricted 
only to those constants that appear in a theory. 

The second part of the rule file contains the program itself, that 
is, a collection of clauses describing constraints of the problem to be
solved. 

By a {\em term tuple} we mean a tuple whose each component is a variable
or a constant symbol, or an arithmetic expression.
An atom is an expression of one of the following four forms.
\begin{enumerate}
\item $p(t)$, where $p$ is a predicate (possibly a predefined
predicate)
and $t$ is a tuple of variables, constants and arithmetic expressions.
\item $p(t,Y):dp(Y)$, where $p$ is a program predicate, $t$ is a term
tuple, and $dp$ is a unary data predicate
\item $m\{p(t):d_1(t_1):\ldots: d_k(t_k)\}n$, where $p$ is a program
predicate, each $d_i$ is a data or a predefined predicate, and $t$ and 
all $t_i$ are term tuples
\item $m\{p_1(t),\ldots,p_k(t)\}n$, where all $p_i$ are program predicates
and $t$ is a term tuple
\end{enumerate}
Atoms of the second type are called {\em e-atoms} and atoms of types 3
and 4 are called {\em c-atoms}. Intuitively, an e-atom `$p(t,Y):dp(Y)$' 
stands for `there exists $Y$ in the extension of the data predicate
$dp$ such that $p(t,Y)$ is true'. An intuitive meaning of a c-atom
`$m\{p(t):d_1(t_1):\ldots: d_k(t_k)\}n$' is: from the set of all atoms
$p(t)$ such that for every $i$, $1\leq i\leq k$, $d_i(t^{p,i})$ is true
($t^{p.i}$ is a projection of $t$ onto attributes of $d_i$),
at least $m$ and no more than $n$ are true. The meaning of a c-atom
`$m\{p_1(t),\ldots,p_k(t)\}n$' is similar: at least $m$ and no more 
than $n$ atoms in the set $\{p_1(t),\ldots,p_k(t)\}$ are true.

We are now ready to define clauses. They are expressions of the form
\[
A_1,\ldots, A_m \rightarrow B_1|\ldots|B_n.
\]
where $A_i$'s and $B_j$'s are atoms, `,' stands for the conjunction
operator and `$|$' stands for the disjunction operator.
 
\section{Processing $\eplc$-theories}
\label{exec}

To compute models of a $\eplc$-theory $(D,P)$ we first ground it. To
this end, we use the program {\em psgrnd}. Next, we compute models of 
the ground theory produced by {\em psgrnd}. To accomplish this task, we
use the program {\em aspps}. For the detailed description of the 
grounding process and, especially, for the treatment of e-atoms and 
c-atoms, and for a discussion of the design of the {\em aspps} program,
we refer the reader to \cite{et01a}.

The required input to execute {\em psgrnd} is a single program file, one
or more data files and optional constants.  If no errors are found while
reading the files and during grounding, an output file is constructed.
The output file is a machine readable file whose name is a catenation of the
constants and file names with the extension {\bf .tdc}. \\
{\bf {\em psgrnd} -r rfile -d dfile1 dfile2 $\ldots$ [-c c1=v1 c2=v2 $\ldots$]}

\smallskip
\noindent
{\bf Required arguments}
\begin{itemize}
\item[-r]
{\bf rfile} is the file describing the problem (rule file). There must
be exactly one rule file.  
\item[-d]
{\bf datafilelist} is one or more
files containing data that will be used to instantiate the theory.
\end{itemize}
{\bf Optional arguments}
\begin{itemize}
\item[-c]
{\bf name=value}
This option allows the use of constants in both the data and rule
files.  When {\bf name} is found while reading input files it is
replaced by {\bf value};  {\bf value} can be any string that is valid
for the data type. If {\bf name} is to be used in a range
specification, then {\bf value} must be an integer.  
\end{itemize}

The program {\em aspps} is used to solve the grounded theory constructed by 
{\em psgrnd}. The name of the file containing the theory is input 
on the command line. After executing the {\em aspps} program, a file 
named aspps.stat is created or appended with statistics concerning this
run of {\em aspps}.

\noindent
{\bf {\em aspps} -f filename  [-A] [-P] [-C [x]] [-S name]}

\smallskip
\noindent
{\bf Required arguments}
\begin{itemize}
\item[-f]
{\bf filename} is the name of the file containing a theory produced by
{\em psgrnd}.
\end{itemize}
{\bf Optional arguments}
\begin{itemize}
\item[-A]
Prints the positive atoms for solved theories in readable form. 
\item[-P]
Prints the input theory and then exits.
\item[-C] {\bf [x]}
Counts the number of solutions. This information is recorded in the statistics
file. If {\bf x} is specified it must be a positive integer; {\em
aspps} stops after finding {\bf x} solutions or exhausting the whole
search space, whichever comes first.
\item[-S] {\bf name} 
Show positive atoms with predicate name. 
\end{itemize}


\newcommand{\etalchar}[1]{$^{#1}$}

\end{document}